\documentclass[a4paper]{cas-dc}

\usepackage[numbers]{natbib}

\usepackage{amsmath,amsfonts}
\usepackage{amssymb}
\usepackage{booktabs}
\usepackage{graphicx}
\usepackage{multirow}
\usepackage[table]{xcolor}
\usepackage{algorithm}
\usepackage{algorithmic}

\usepackage{afterpage}
\usepackage{enumitem}

\hypersetup{
    colorlinks=true,
    citecolor=blue,
    linkcolor=blue,
}

\newcommand{\best}[1]{\textbf{\textcolor{red}{#1}}}
\newcommand{\second}[1]{\textbf{\textcolor{blue}{#1}}}
\newcommand{\tabhead}[1]{\begin{tabular}[c]{@{}c@{}}#1\end{tabular}}



\begin{document}

\shorttitle{AIGS-Net}
\shortauthors{Yuhan Chen et~al.}

\title [mode = title]{AIGS-Net: Compact Illumination Field Modeling via 2D Gaussian Splatting for Fast Low-Light Image Enhancement}

\author[1]{Yuhan Chen}
\ead{20240701028@stu.cqu.edu.cn}

\author[2]{Kunyang Huang}
\ead{kunyangh@andrew.cmu.edu}

\author[3]{Fuchen Li}
\ead{fuchen.li@ufl.edu}

\author[4]{Zhuohan Qin}
\ead{qinzhuohan@qdu.edu.cn}

\author[1]{Guofa Li}\cormark[1]
\ead{liguofa@cqu.edu.cn}

\author[5]{Wenbo Chu}
\ead{chuwenbo@wicv.cn}

\author[6]{Keqiang Li}
\ead{likq@tsinghua.edu.cn}

\affiliation[1]{organization={College of Mechanical and Vehicle Engineering, Chongqing University},
                city={Chongqing},
                postcode={400044},
                country={China}}

\affiliation[2]{organization={Department of Electrical and Computer Engineering, Carnegie Mellon University},
                city={Moffett Field},
                postcode={CA 94035},
                country={USA}}

\affiliation[3]{organization={Herbert Wertheim College of Engineering, University of Florida},
                city={Gainesville},
                postcode={FL 32611},
                country={USA}}

\affiliation[4]{organization={School of Mathematics and Statistics, Qingdao University},
                city={Qingdao},
                country={China}}

\affiliation[5]{organization={National Innovation Center of Intelligent and Connected Vehicles},
                city={Beijing},
                postcode={100089},
                country={China}}

\affiliation[6]{organization={School of Vehicle and Mobility, Tsinghua University},
                city={Beijing},
                postcode={100084},
                country={China}}

\cortext[cor1]{Corresponding author.}

\begin{abstract}
Existing low-light image enhancement methods often face a bottleneck between the representation capacity of illumination-field modeling and computational complexity. To address this issue, this paper proposes an Adaptive Illumination Gaussian Splatting Network (AIGS-Net), an ultra-lightweight architecture for fast low-light enhancement. Unlike conventional static priors, AIGS-Net constructs an input-adaptive 2D Gaussian Splatting illumination field. The opacity of Gaussian basis functions is dynamically modulated by relative luminance statistics of the input image, and spatially varying illumination compensation is rendered through ordered alpha compositing. To guide adaptive illumination gain efficiently, a zero-parameter nonlinear multiscale contextual encoding module is introduced to extract low-frequency structures and local contrast cues without additional convolutional weights. To suppress noise amplification and sensor-induced color bias, AIGS-Net integrates noise-mask estimation, locked single-channel Gamma mapping, cross-channel consistency regularization, and target color-alignment constraints. Experiments on LOL and LSRW benchmarks show that AIGS-Net improves detail recovery and color fidelity while requiring only approximately 440 learnable parameters, achieving an effective trade-off between enhancement quality and extreme inference efficiency.
\end{abstract}

\begin{highlights}
\item AIGS-Net builds input-adaptive 2DGS illumination fields.
\item Relative luminance statistics modulate Gaussian opacity.
\item Zero-parameter multiscale context guides structure-aware gain.
\item Locked Gamma and noise masks suppress color shift and noise.
\item AIGS-Net achieves strong quality with only about 440 parameters.
\end{highlights}


\maketitle

\section{Introduction}
Images captured under low-light conditions often exhibit severe visibility degradation, reduced contrast, prominent noise in dark regions, and sensor-induced color bias~\cite{ref1,ref2}. These degradations affect human visual perception and severely limit the performance of high-level downstream computer vision tasks, such as object detection and autonomous driving~\cite{ref3}. Therefore, low-light image enhancement (LLIE) has remained a fundamental problem in image processing and computer vision.

Early LLIE methods were mainly based on histogram equalization or Retinex-based physical models~\cite{ref26}, where enhancement is performed by imposing prior assumptions on reflectance and illumination maps. In recent years, deep learning has advanced this field significantly. Visual quality has been substantially improved by lightweight methods based on zero-reference deep curve estimation~\cite{ref6,ref7}, residual networks with complex attention mechanisms and multiscale feature fusion~\cite{ref11,ref12}, and recent generative paradigms based on diffusion models~\cite{ref21,ref22} and implicit neural representations~\cite{ref24,ref25}. However, these advanced models often face a difficult bottleneck between the representation capacity of illumination-field modeling and computational complexity. For example, diffusion models and large implicit neural representation networks can generate high-quality images, but their large parameter sizes and high inference latency make them difficult to deploy on resource-constrained mobile devices or edge vision systems with strict real-time requirements~\cite{ref4,ref20}.

Recently, 3D Gaussian Splatting (3DGS) has attracted considerable attention due to its excellent performance in 3D scene reconstruction and real-time rendering~\cite{ref28,ref29}. Two-dimensional Gaussian Splatting (2DGS) has since been introduced into image processing and has demonstrated high computational and parameter efficiency in tasks such as image representation~\cite{ref51}, image compression~\cite{ref50}, and super-resolution~\cite{ref54}. Although preliminary attempts have applied 2DGS to low-light enhancement~\cite{ref47,ref48}, most existing methods model the illumination field as a fixed mapping over static Gaussian bases, while ignoring the substantial differences in spatial luminance distributions across input images. When low-light scenes exhibit substantially different brightness distributions, the illumination field may degrade, leading to inaccurate spatially varying illumination compensation. In addition, existing Gaussian Splatting image models usually lack explicit modeling of low-light-induced noise and multichannel relative white balance, which can amplify noise and preserve color bias during brightening.

To overcome the bottleneck between representation capacity and computational efficiency, this paper proposes an ultra-lightweight Adaptive Illumination Gaussian Splatting Network (AIGS-Net). Unlike conventional static priors, AIGS-Net constructs an input-adaptive 2DGS illumination field. Specifically, relative luminance statistics are extracted from the input image to dynamically modulate the opacity of each Gaussian basis function. An illumination compensation field that accurately adapts to the current scene structure is then rendered through alpha depth compositing. To efficiently guide illumination gain without introducing additional parameters, a nonlinear multiscale contextual encoding module is designed. This module extracts local mean and local contrast features purely from the spatial domain through shift-based aggregation. To suppress noise and color distortion caused by high gain, AIGS-Net introduces noise-mask estimation and locked single-channel Gamma tone mapping, supported by cross-channel consistency and target color-angle losses. This design reduces the tendency through which low-light color bias is preserved by the network.

In summary, the main contributions of this paper are as follows:
\begin{enumerate}[leftmargin=*,label=\arabic*)]
\item An ultra-lightweight AIGS-Net architecture with an input-adaptive 2DGS illumination field is proposed. The limitation of conventional static illumination mapping is overcome by dynamically modulating Gaussian opacity with relative luminance statistics. Accurate spatially varying illumination compensation is achieved through alpha compositing, and the entire network requires only approximately 440 learnable parameters.
\item A zero-parameter nonlinear multiscale contextual encoding module is designed. Without introducing any learnable convolutional weights, low-frequency structures and high-frequency local contrast features are efficiently extracted through multiscale spatial shift operations, which provide structure-aware gating guidance for adaptive illumination gain.
\item A complete physical degradation restoration and consistency constraint mechanism is constructed. For low-light-induced noise and sensor color bias, noise estimation and single-channel Gamma mapping are integrated. Cross-channel gain-variance regularization and a target color-alignment loss are further imposed. Experimental results show that AIGS-Net achieves fast inference while providing better detail recovery and color fidelity, reaching a favorable trade-off between enhancement quality and extreme efficiency.
\end{enumerate}

\section{Related Work}
This section first reviews recent advances in deep learning for low-light image enhancement. It then discusses the evolution of Gaussian Splatting techniques in visual rendering and image representation. Finally, the limitations of existing Gaussian-based enhancement models are analyzed in detail, followed by a detailed description of the solution provided by AIGS-Net.

\subsection{Deep Learning for Low-Light Image Enhancement}
Low-light image enhancement aims to recover high-quality, clear images from degraded observations affected by insufficient illumination, noise, and color distortion~\cite{ref1}. With continuous technical progress over the past few years, deep learning methods have become the dominant paradigm in this field~\cite{ref2,ref3}. Early classical methods were typically based on illumination-map estimation or Retinex-based physical models, such as LIME~\cite{ref5} and Retinex-Net~\cite{ref26}. Enhancement was achieved by explicitly decomposing an image into reflectance and illumination maps. With the evolution of network architectures, several efficient paradigms have been developed, including the Zero-DCE series based on zero-reference deep curve estimation~\cite{ref6,ref7}, its curve-optimization extension ChebyLighter~\cite{ref14}, networks that combine architecture search with physical model unfolding~\cite{ref9,ref13,ref18}, and multiscale residual networks designed to improve inference speed~\cite{ref11,ref12}. To balance performance and efficiency, the field has further explored fast and flexible robust enhancement architectures~\cite{ref8}, R2RNet for mapping real low-light images to normal-light images~\cite{ref27}, EFINet for image restoration through iterative enhancement and fusion~\cite{ref15}, and minimalist enhancement networks learned from paired low-light instances~\cite{ref16}. Meanwhile, unsupervised learning~\cite{ref10}, noise autoregression~\cite{ref17}, and zero-shot illumination-guided models~\cite{ref19} have also been proposed to improve generalization in real-world scenarios.

In recent years, generative artificial intelligence has introduced new paradigms for LLIE. Diffusion models~\cite{ref21,ref22,ref23}, implicit neural representations including SIREN~\cite{ref31} and its derived models~\cite{ref24,ref25}, and state-space architectures~\cite{ref55} have achieved high visual quality in detail recovery and global mapping. However, these models incur extremely high computational costs due to cumbersome iterative denoising procedures or large parameter scales, which makes deployment almost infeasible in edge vision systems with strict requirements for real-time processing and ultra-low power consumption~\cite{ref4,ref20}. Therefore, designing an ultra-lightweight enhancement architecture that does not sacrifice physical modeling capacity remains an urgent challenge.

\subsection{Gaussian Splatting in Vision and Rendering}
Three-dimensional Gaussian Splatting was originally developed to address the slow rendering speed of Neural Radiance Fields~\cite{ref28,ref29}. By representing a scene with anisotropic 3D Gaussian ellipsoids and combining this representation with rasterization, 3DGS enables real-time rendering with high visual quality. To further improve reconstruction and rendering efficiency, fast and generalizable reconstruction schemes, such as Speedy-Splat~\cite{ref33} and MVSGaussian~\cite{ref34}, have been proposed, followed by GaussianPro based on progressive propagation~\cite{ref39} and momentum Gaussian self-distillation strategies~\cite{ref41}. Owing to its strong representation capacity and computational efficiency, 3DGS has been rapidly extended to a wide range of tasks, including large-scale urban scene and periodic vibration scene rendering~\cite{ref32,ref37,ref40}, reconstruction of complex dynamic autonomous driving environments~\cite{ref38,ref42,ref43}, long-span spatiotemporal motion modeling~\cite{ref35,ref36}, text generation that bridges 2D and 3D diffusion models~\cite{ref44}, and skeleton-based action recognition~\cite{ref56}.

To adapt Gaussian Splatting to two-dimensional manifolds, 2DGS was subsequently developed~\cite{ref30}. 2DGS directly optimizes the mean, covariance, and opacity of Gaussian bases on the image plane, showing remarkable potential in purely 2D image processing. Representative examples include GaussianImage, which first achieved thousand-frame-level image representation and compression~\cite{ref51}; Instant GaussianImage, which further improved generalization and adaptive representation efficiency~\cite{ref50}; efficient dataset distillation based on sparse Gaussian representation~\cite{ref52}; large-scale high-quality image fitting~\cite{ref49}; vision-language feature alignment~\cite{ref53}; and arbitrary-scale image super-resolution~\cite{ref54}. These pioneering studies demonstrate the compactness of 2DGS in continuous 2D signal representation and provide a new tool for overcoming the pixel-wise mapping bottleneck of conventional CNNs.

\begin{figure*}[!t]
\centering
\includegraphics[width=1.0\textwidth]{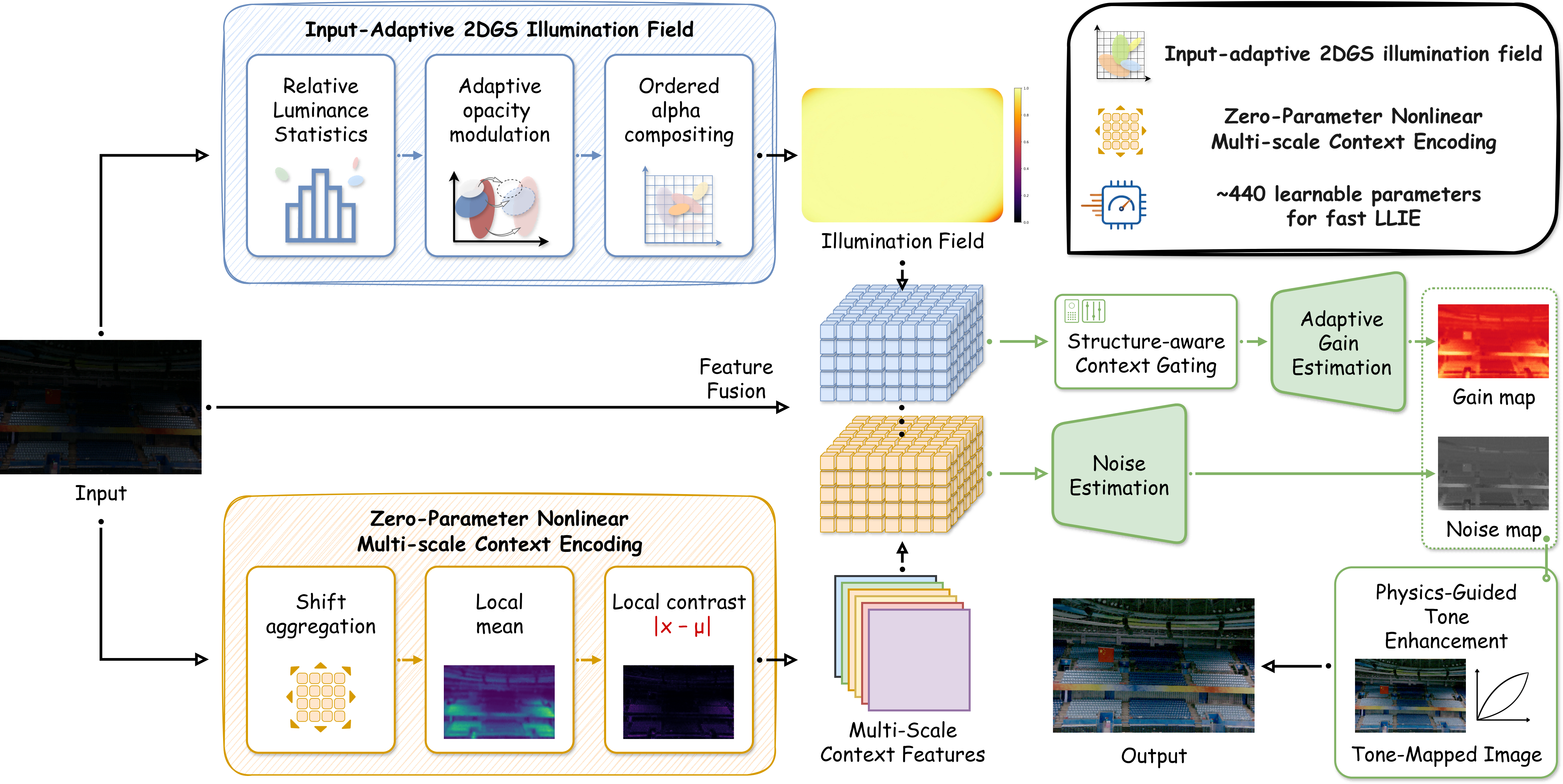}
\caption{The overall architecture of the proposed AIGS-Net. For a low-light input image, AIGS-Net first constructs an illumination field via the input-adaptive 2DGS branch. This branch extracts relative luminance statistics from the image, dynamically modulates the opacity of Gaussian basis functions, and renders an input-dependent illumination prior through ordered alpha compositing. Meanwhile, the zero-parameter nonlinear multiscale contextual encoding branch extracts local mean and contrast features via shift aggregation, providing structure-aware guidance without additional learnable convolutional kernels. The illumination field and the input image are then fused to estimate the adaptive gain and noise maps, with multiscale contextual features contributing to gain prediction via structure-aware gating. Finally, the physically guided tone-enhancement module jointly performs gain compensation, noise suppression, single-channel Gamma mapping, and residual fusion to generate the final enhanced result.}
\label{fig:framework}
\end{figure*}

\subsection{2DGS for Image Enhancement and Limitations}
\sloppy Inspired by the strong generalization capability of Gaussian Splatting, recent studies have begun to introduce it into low-light reconstruction and enhancement. From the 3D perspective, LL-Gaussian~\cite{ref45} and LLGS~\cite{ref46} attempted novel-view synthesis and reconstruction under low-light conditions. For LLIE tasks based on 2DGS image-compression-domain representation, LL-GaussianImage~\cite{ref47} first achieved low-light image enhancement in the 2DGS image compression domain, while LL-GaussianMap~\cite{ref48} first used 2DGS priors to generate gain maps for unsupervised low-light image enhancement.

Nevertheless, existing Gaussian-based enhancement methods still suffer from critical limitations in illumination modeling and degradation restoration. First, current 2DGS-based LLIE methods mainly focus on compression-domain enhancement or structure-prior gain maps. The modeling of input-dependent spatial luminance distributions, noise propagation, and cross-channel color-cast constraints remains insufficient. These methods ignore substantial differences in spatial luminance distributions among input images, such as scenes that are bright on the left and dark on the right or bright in the center and dark near the boundaries. When low-light scenes exhibit highly disparate brightness distributions, static priors can severely degrade and fail to provide accurate spatially varying compensation. Second, existing Gaussian models lack explicit constraints on low-light-induced noise and multichannel color distortion. Therefore, high-frequency noise can be easily amplified during brightening, and sensor-induced color bias can be preserved.

To overcome these bottlenecks, AIGS-Net introduces effective innovations from three aspects. First, to address the degradation of static priors, AIGS-Net constructs an input-adaptive 2DGS illumination field. Relative luminance statistics are extracted from the image to dynamically modulate the opacity of Gaussian bases, and an illumination-compensation field that accurately matches the current scene structure is rendered via alpha-depth compositing. Second, to efficiently guide adaptive gain, the model designs a zero-parameter nonlinear multiscale contextual encoding module. Local mean and local contrast features are extracted solely via spatial shifts, avoiding redundant convolutional parameters. Meanwhile, to address degradation amplification, AIGS-Net introduces noise-mask estimation and forcibly constrains the image to a single-channel Gamma tone-curve mapping. Cross-channel gain variance regularization and target color-alignment loss are jointly imposed. Finally, with only approximately 440 learnable parameters, AIGS-Net reduces the tendency through which low-light color bias is preserved by the network under an extremely lightweight setting, achieving a favorable trade-off between visual quality and inference efficiency.

\section{Proposed Method}
This paper proposes AIGS-Net, an ultra-lightweight network for fast low-light image enhancement. The core idea is not to rely on stacked deep convolutional networks 
\begin{figure}
\centering
\includegraphics[width=\columnwidth]{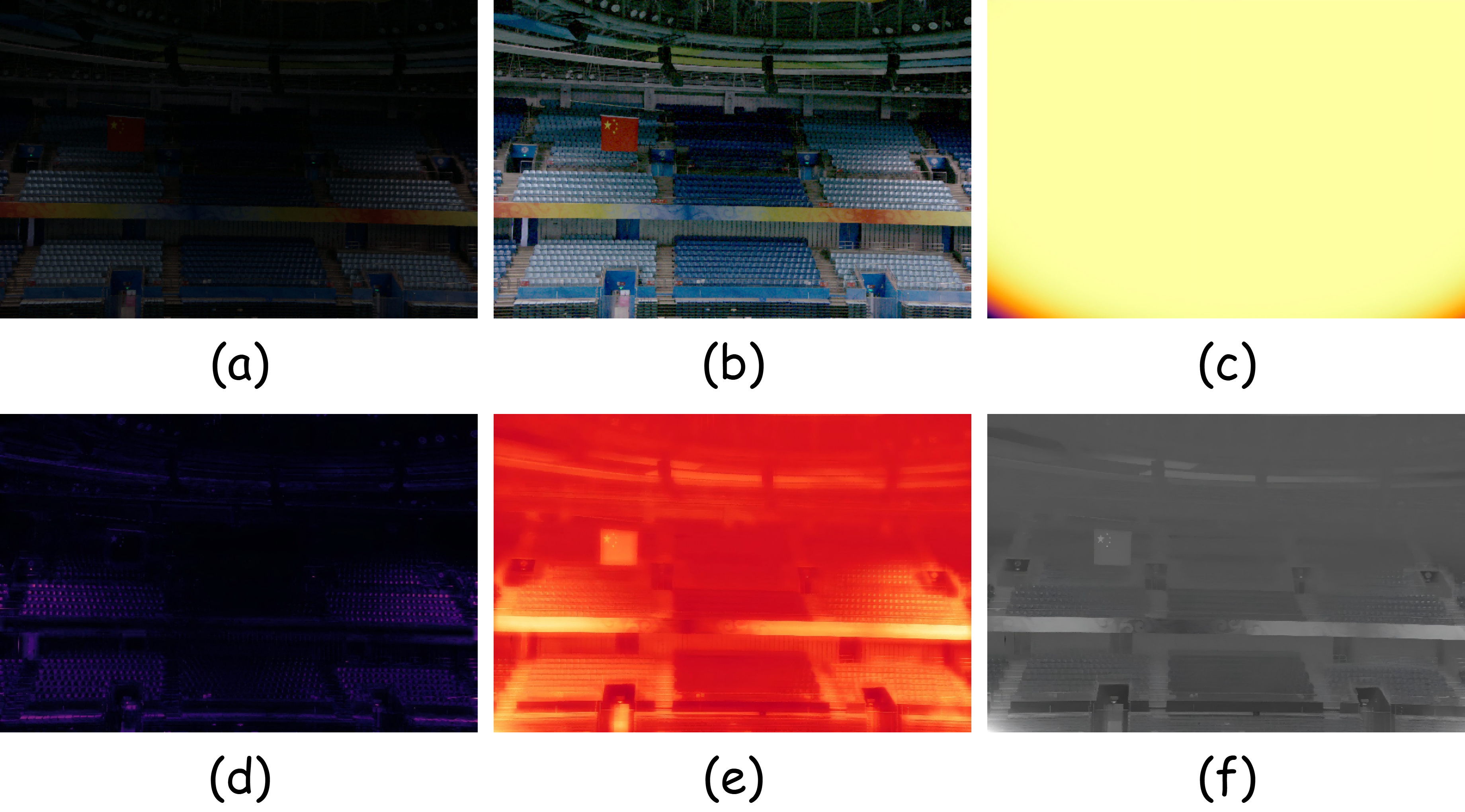}
\caption{Visualization of physical prior decomposition and enhancement in AIGS-Net. From left to right, the images are arranged as input, output, illumination field, gain field, and noise map.}
\label{fig:prior_decomposition}
\end{figure}
to learn a complex end-to-end mapping. Instead, the low-light enhancement process is explicitly decomposed into three physically meaningful subproblems: how to represent input-dependent spatial illumination distributions with very few parameters, how to introduce local structural and textural context without significantly increasing computational cost, and how to suppress noise amplification and color drift during high-gain brightening. To this end, AIGS-Net integrates an input-adaptive 2DGS illumination field, zero-parameter nonlinear multiscale contextual encoding, and a physically guided tone-enhancement process into a compact framework.

Given a low-light image $I_l \in [0,1]^{3\times H\times W}$, the network jointly predicts an adaptive illumination field $L$, a channel-wise gain field $\Gamma$, and a noise estimation map $\epsilon$, and then generates the enhanced result $\hat{I}$. The forward mapping of the overall model can be summarized as follows:
\begin{equation}
\begin{gathered}
(L,\Gamma,\epsilon,\hat{I}) = \mathcal{F}_{\Theta}(I_l),\\
\Theta = \{\Theta_{\mathrm{2DGS}},\Theta_{1\times 1},\gamma_0,r_0\},
|\Theta| \approx 440 .
\end{gathered}
\label{eq:overall_forward}
\end{equation}
where $\Theta_{\mathrm{2DGS}}$ denotes the center, shape, opacity, and adaptive modulation parameters in the 2DGS illumination field; $\Theta_{1\times 1}$ denotes a small number of pointwise convolution parameters; $\gamma_0$ is the global single-channel Gamma tone parameter; and $r_0$ is the residual fusion weight. Under the default setting of $K=64$, the 2DGS branch contains only $6K+2$ learnable parameters. The subsequent pointwise convolutions and global scalar parameters bring the whole network to approximately 440 parameters. This design preserves the representation capacity for spatially non-uniform illumination while avoiding the storage and inference burden caused by numerous convolution kernels in conventional deep enhancement networks.

The overall pipeline is shown in Fig.~\ref{fig:framework}. The low-light input is first fed into the input-adaptive 2DGS illumination-field branch. The network dynamically modulates the opacity of each Gaussian basis function based on the relative luminance statistics of the current image and produces a spatially continuous illumination prior via alpha compositing. Meanwhile, the input image is processed by zero-parameter multiscale shift aggregation to obtain local mean and local contrast features. The former reflects low-frequency luminance structures, whereas the latter captures edges, textures, and local variations. The illumination field, original input, and contextual features are then jointly used to guide the estimation of the gain field and noise map. Finally, the enhanced image is generated through gain compensation, noise subtraction, single-channel Gamma tone mapping, and weak residual fusion. Unlike purely data-driven deep networks, each intermediate variable in AIGS-Net has a clear physical or structural interpretation. To demonstrate the interpretability of AIGS-Net, Fig.~\ref{fig:prior_decomposition} presents the intermediate physical priors of a typical sample.

\subsection{Input-Adaptive 2D Gaussian Splatting Illumination Field}
The key to low-light image enhancement lies in estimating spatially non-uniform illumination degradation. Conventional Retinex-based methods typically produce a smooth illumination map via filtering or optimization, whereas many deep learning methods implicitly learn illumination compensation from convolutional features. However, the former has limited representation capacity, while the latter incurs a high computational cost. In this paper, the illumination field is modeled by 2D Gaussian Splatting, where a limited number of learnable anisotropic 2D Gaussian basis functions are used to represent spatial illumination structures. Different from static Gaussian bases, the core improvement of AIGS-Net is that the opacity of each Gaussian basis function is not a fixed parameter but is dynamically modulated by the relative luminance distribution of the current input image. Therefore, the same set of Gaussian basis functions can render different illumination compensation fields for different low-light images.

First, the input image is converted into a luminance map $Y$. The luminance component is obtained using the standard RGB weighted form:
\begin{equation}
Y = 0.299 I_l^R + 0.587 I_l^G + 0.114 I_l^B .
\label{eq:luminance}
\end{equation}

For the $k$-th Gaussian basis function, its center is denoted as $\mu_k=(\mu_{kx},\mu_{ky})$, which is defined in the normalized coordinate space $[-1,1]^2$. In the implementation, the Gaussian centers are initialized on a uniform grid within $[-0.8,0.8]$, which ensures that the illumination basis functions cover the entire image plane at the initial training stage. The Gaussian shape is controlled by a symmetric positive-definite precision matrix:
\begin{equation}
\begin{gathered}
\mathbf{P}_k =
\begin{bmatrix}
a_k & b_k\\
b_k & c_k
\end{bmatrix},
  a_k=\mathrm{Softplus}(\tilde{a}_k)+0.5,\qquad \\
b_k=\tanh(\tilde{b}_k)\frac{\sqrt{a_kc_k}}{2} ,
  c_k=\mathrm{Softplus}(\tilde{c}_k)+0.5 .
\end{gathered}
\label{eq:precision_matrix}
\end{equation}
At coordinate $p=(x,y)$, the response of the $k$-th 2D Gaussian basis function is defined as
\begin{equation}
\begin{gathered}
G_k(p)=\exp\!\left[-\frac{1}{2}(p-\mu_k)^\mathrm{T}\mathbf{P}_k(p-\mu_k)\right],  
p\in[-1,1]^2 .
\end{gathered}
\label{eq:gaussian_response}
\end{equation}
The above formulation allows each Gaussian basis function to learn different scales, orientations, and spatial coverage ranges. Unlike regular convolution kernels, Gaussian Splatting basis functions operate directly in continuous coordinate space, leading to higher parameter efficiency. For low-light enhancement, this continuous spatial representation is particularly suitable for modeling large-scale, slowly varying, and spatially non-uniform illumination degradation.

Using fixed Gaussian responses alone is still insufficient to handle differences in luminance distributions across input images. To avoid degeneration into a static prior, AIGS-Net samples the input luminance at each Gaussian center and computes the relative darkness with respect to the global luminance statistics of the current image. Specifically, the local luminance value at the Gaussian center $\mu_k$ is obtained by bilinear sampling, while the luminance mean and standard deviation of the whole image are computed as follows:
\begin{equation}
\begin{gathered}
y_k(I_l)=\operatorname{GridSample}(Y,\mu_k),\\
\bar{Y}=\frac{1}{|\Omega|}\sum_{p\in\Omega}Y(p),
\sigma_Y=\sqrt{\frac{1}{|\Omega|}\sum_{p\in\Omega}\left(Y(p)-\bar{Y}\right)^2} .
\end{gathered}
\label{eq:relative_stats}
\end{equation}
Then, the normalized relative darkness of the $k$-th Gaussian center is defined as
\begin{equation}
d_k(I_l)=\frac{\bar{Y}-y_k(I_l)}{\max(\sigma_Y,10^{-3})} .
\label{eq:relative_darkness}
\end{equation}
When $d_k(I_l)>0$, the location of the Gaussian center is darker than the average luminance of the current image and should receive stronger illumination compensation. When $d_k(I_l)<0$, this location is relatively bright and should be assigned a lower compensation strength. Unlike the direct use of absolute luminance, relative darkness can effectively distinguish spatial distribution patterns.

The base opacity of each Gaussian basis function is obtained by applying a sigmoid mapping to the learnable parameter $\alpha_k$ and is further dynamically modulated by relative darkness. In the implementation, at least 20\% of the base response is retained to prevent certain Gaussian basis functions from being completely deactivated at the early training stage:
\begin{equation}
\hat{\alpha}_k(I_l)=\sigma(\alpha_k)\left[0.2+0.8\sigma\left(sd_k(I_l)+\beta\right)\right],
\label{eq:adaptive_alpha}
\end{equation}
where $s$ and $\beta$ are learnable scalars. This design enables the Gaussian opacity to be both globally learnable and input-adaptive.

To simulate occlusion and accumulation effects in Gaussian Splatting, front-to-back alpha compositing is further introduced. In the implementation, a stable compositing order is constructed using the distance from each Gaussian center to the coordinate origin:
\begin{equation}
\pi=\operatorname{argsort}\left(\|\mu_k\|_2^2\right) .
\label{eq:sort}
\end{equation}

For the $m$-th Gaussian basis function after sorting, its transmittance is obtained by accumulating the remaining transparency of all preceding Gaussian basis functions:
\begin{equation}
T_m(p;I_l)=\prod_{j<m}\left(1-\hat{\alpha}_{\pi_j}(I_l)G_{\pi_j}(p)\right) .
\label{eq:transmittance}
\end{equation}
Finally, the unnormalized input-adaptive illumination field can be written as
\begin{equation}
\tilde{L}(p;I_l)=\sum_{m=1}^{K}T_m(p;I_l)\hat{\alpha}_{\pi_m}(I_l)G_{\pi_m}(p) .
\label{eq:illumination_unnormalized}
\end{equation}
In the implementation, a small-value clipping operation is applied during transmittance updating to ensure numerical stability. Since $L$ is expected to mainly serve as a spatial structural prior rather than directly represent absolute exposure, the illumination field of each image is independently normalized:
\begin{equation}
L(p;I_l)=\frac{\tilde{L}(p;I_l)-\min_{q\in\Omega}\tilde{L}(q;I_l)}{\max_{q\in\Omega}\tilde{L}(q;I_l)-\min_{q\in\Omega}\tilde{L}(q;I_l)+10^{-8}} .
\label{eq:illumination_norm}
\end{equation}
This normalization constrains the illumination field to $[0,1]$ and highlights the relative illumination distribution of the current input. The subsequent gain estimation branch then determines the actual brightening magnitude based on $I_l$, $L$, and contextual features.

\begin{figure*}[!t]
\centering
\includegraphics[width=1.0\textwidth]{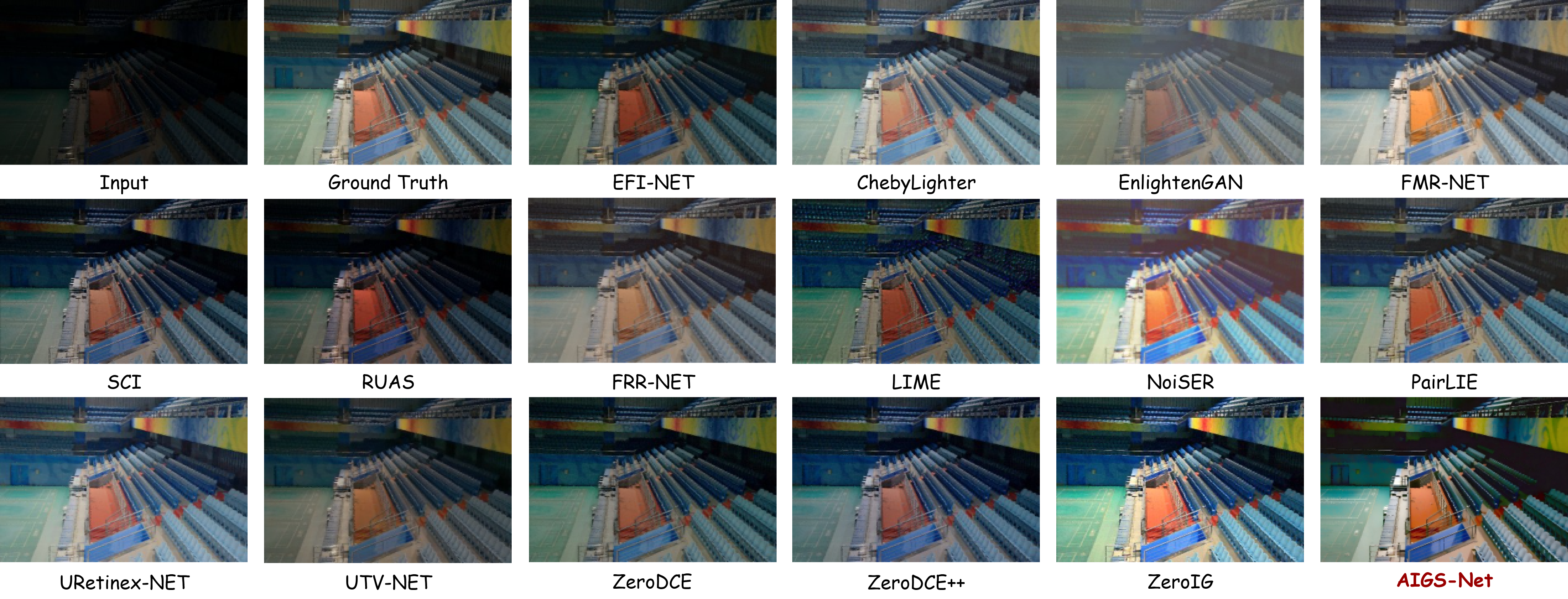}
\caption{Visual comparison between AIGS-Net and SOTA methods on the LOL dataset.}
\label{fig:lol_visual}
\end{figure*}

\begin{figure*}[!t]
\centering
\includegraphics[width=1.0\textwidth]{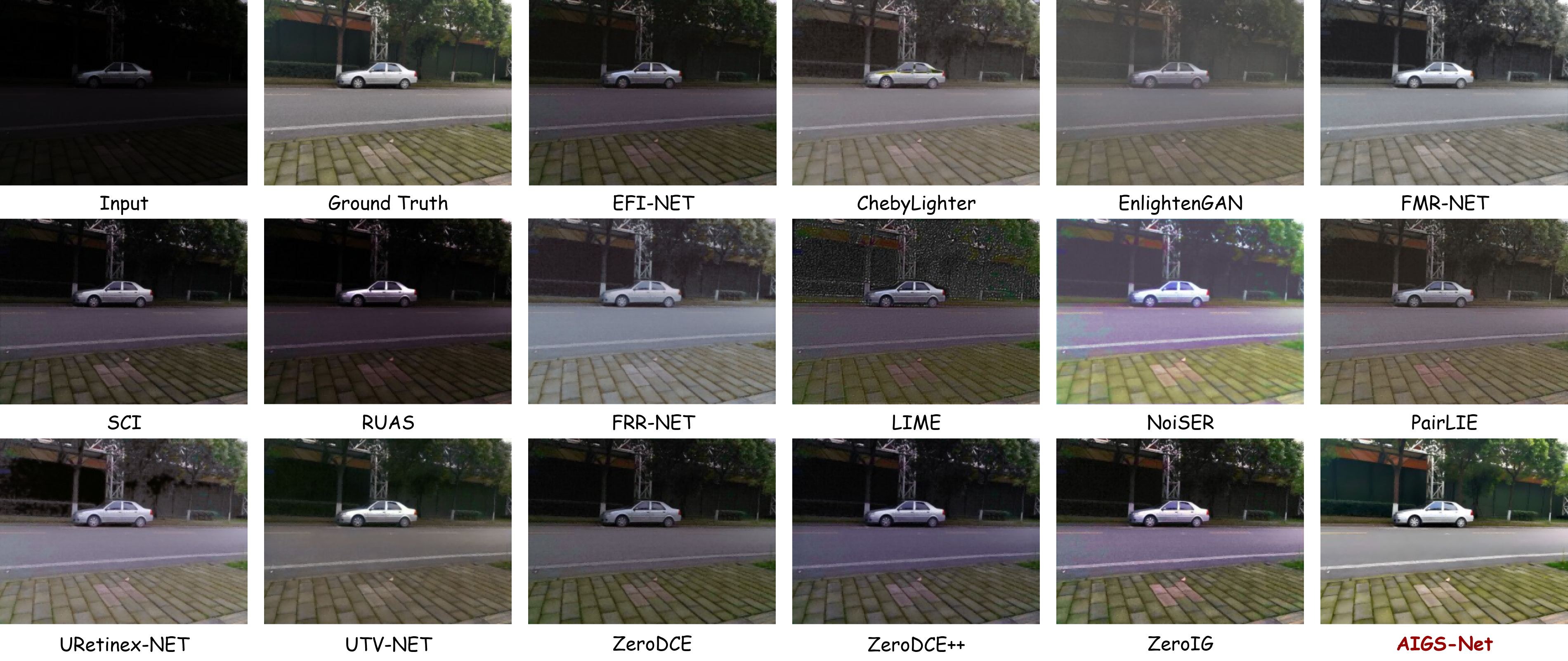}
\caption{Visual comparison between AIGS-Net and SOTA methods on the LSRW-HUAWEI dataset.}
\label{fig:huawei_visual}
\end{figure*}

\subsection{Zero-Parameter Nonlinear Multiscale Contextual Encoding}
Relying solely on the illumination field $L$ can capture large-scale luminance distributions, but low-light enhancement also requires identifying edges, textures, and local contrast variations to avoid over-smoothing or local overexposure. A common practice is to introduce additional convolutional layers, attention modules, or multiscale feature pyramids, but this significantly increases the number of parameters and computational complexity. To maintain extreme lightweightness, AIGS-Net designs a zero-parameter nonlinear multiscale contextual encoding module. This module does not contain any learnable convolution kernels and obtains local mean and local contrast features only through spatial shifting and summation.

Given a dilation rate $r$, shift aggregation is first performed on the input image within a $3\times3$ neighborhood. In the implementation, zero padding is used, and neighboring pixels are sampled spatially according to the dilation interval $r$:
\begin{equation}
S_r(I_l)(p,c)=\sum_{u=-1}^{1}\sum_{v=-1}^{1} I_l\bigl(p+r(u,v),c\bigr) .
\label{eq:shift_aggregation}
\end{equation}
The corresponding local mean is given by
\begin{equation}
M_r(I_l)=\frac{1}{9}S_r(I_l) .
\label{eq:local_mean}
\end{equation}
In this paper, three scales are used by default, with $\mathcal{R}=\{3,5,7\}$. The multiscale local mean is obtained by averaging the local means across all scales, while the multiscale local contrast is obtained by averaging the absolute differences between the input and the local mean at each scale:
\begin{equation}
\begin{gathered}
C_m=\frac{1}{|\mathcal{R}|}\sum_{r\in\mathcal{R}}M_r(I_l),\\
C_c=\frac{1}{|\mathcal{R}|}\sum_{r\in\mathcal{R}}\left|I_l-M_r(I_l)\right|,
\mathcal{R}=\{3,5,7\} .
\end{gathered}
\label{eq:contextual_encoding}
\end{equation}
where $C_m$ mainly describes low-frequency spatial structures and local background luminance, while $C_c$ characterizes nonlinear local contrast, edge, and texture responses. The absolute-difference operation in $C_c$ yields contextual encoding that is more than simple linear smoothing, enabling it to explicitly respond to local structural changes. This is particularly important for low-light enhancement, because details in dark regions usually appear as weak local contrast on low-luminance backgrounds.

The contextual encoding itself introduces no learnable parameters. To transform it into a structural guidance term for gain prediction, AIGS-Net uses only two sets of channel-wise $1\times1$ mappings. Both mappings are implemented as depthwise channel-wise convolutions with $\mathrm{Groups}=3$, where each color channel is adjusted independently:
\begin{equation}
C_s=\sigma\left(\phi_g(C_c)\right)\odot\operatorname{ReLU}\left(\phi_v(C_m)\right) .
\label{eq:context_gate}
\end{equation}
where $\phi_g$ generates the structure-aware gate, and $\phi_v$ generates the contextual response based on the local mean. This design has two advantages. First, the gate is controlled by local contrast $C_c$, allowing edge and texture regions to have a stronger influence on gain estimation. Second, since both $\phi_g$ and $\phi_v$ are channel-wise pointwise convolutions, they require very few parameters and do not compromise the lightweight design objective.

\subsection{Physics-Guided Tone Enhancement}
After obtaining the input-adaptive illumination field $L$ and multiscale contextual features, AIGS-Net generates the final output through a minimal physics-guided tone-enhancement module. This part consists of a sequence of intermediate processes, including gain estimation, noise estimation, Gamma tone mapping, and residual fusion. Its objective is not only to improve brightness but also to suppress noise amplification and maintain color stability under high-gain conditions.

First, the original low-light input $I_l$ and the normalized illumination field $L$ are concatenated along the channel dimension to form a four-channel physical feature. This feature is passed through a $1\times1$ convolution and a ReLU activation to obtain a compact representation:
\begin{equation}
F=\operatorname{ReLU}\left(\phi_f([I_l,L])\right),
\label{eq:fused_feature}
\end{equation}
where $\phi_f$ denotes a pointwise convolution. This branch only performs cross-channel information fusion and does not introduce additional spatial convolution cost. The network then uses this fused feature together with the contextual guidance term $C_s$ to estimate the channel-wise gain field:
\begin{equation}
\Gamma=\operatorname{clip}\left(1+\operatorname{ReLU}\left(\phi_{\Gamma}(F)+C_s\right),1,12\right) .
\label{eq:gain}
\end{equation}
where $\phi_{\Gamma}$ denotes a pointwise convolution. The gain field $\Gamma\in\mathbb{R}^{3\times H\times W}$ is spatially varying and channel-dependent. The lower bound of 1 ensures the network does not further darken low-light images, while the upper bound of 12 provides sufficient brightening capacity for severely low-light scenes and avoids numerical instability from excessively large values.

Another key issue in low-light enhancement is noise. Noise in dark regions of low-light images is often further amplified after gain enhancement. To address this issue, AIGS-Net estimates a single-channel noise map from the same fused feature $F$:
\begin{equation}
\epsilon=\sigma\left(\phi_{\epsilon}(F)\right) .
\label{eq:noise_map}
\end{equation}
The noise map is constrained to $[0,1]$ and is then subtracted from the gain-compensated image with a small coefficient:
\begin{equation}
J=I_l\odot\Gamma-0.05\epsilon .
\label{eq:noise_suppression}
\end{equation}
where $J$ denotes the physically enhanced image before tone mapping. The coefficient 0.05 controls the strength of noise subtraction and prevents excessive denoising from causing texture loss.

Next, AIGS-Net further adjusts image brightness using global Gamma tone mapping. Unlike common channel-wise Gamma correction, this paper adopts a single shared Gamma parameter, where the RGB channels use the same tone curve. This design is important for color stability. Low-light images often suffer from channel imbalance caused by sensor-induced yellowish or bluish color casts. If each channel is allowed to learn an independent Gamma value, the network may further preserve or even amplify color bias through different nonlinear curves across channels. In contrast, single-channel Gamma correction maintains the relative RGB proportions without additional distortion, which is more favorable for white-balance stability:
\begin{equation}
\gamma=\mathrm{Softplus}(\gamma_0)+0.5, \quad \gamma\in\mathbb{R}^{1\times1\times1} .
\label{eq:gamma}
\end{equation}
Before tone mapping, $J$ is clipped to a stable range:
\begin{equation}
O=\operatorname{clip}(J,10^{-6},1)^{1/\gamma} .
\label{eq:tone_mapping}
\end{equation}
Finally, AIGS-Net introduces a global learnable residual weight to lightly fuse the tone-mapped result with the original input:
\begin{equation}
r=\sigma(r_0), \quad \hat{I}=\operatorname{clip}\left((1-r)O+rI_l,0,1\right) .
\label{eq:residual}
\end{equation}
where $r_0$ is set to $-4$, so the initial residual weight is approximately 0.018. This means that at the early training stage, the model mainly relies on the physical enhancement branch output rather than simply preserving a large portion of the dark input, thereby avoiding insufficient brightness in the enhanced result. As training proceeds, the residual weight can be adaptively adjusted to preserve part of the original structure and suppress over-enhancement.

As shown in Fig.~\ref{fig:huawei_visual}, the enhancement results also differ significantly in the low-light road scene. EFI-NET, SCI, RUAS, and ZeroDCE still produce dark outputs, with vehicle and tree details not sufficiently restored. ChebyLighter, EnlightenGAN, and FRR-NET suffer from haze, whitening, or contrast degradation. LIME, NoiSER, and ZeroIG introduce noise, color casts, and overexposure. In contrast, AIGS-Net achieves moderate brightness enhancement, clearer road textures, sharper vehicle contours, and better background structures, while maintaining natural colors closer to the GT image.

\begin{figure*}[!t]
\centering
\includegraphics[width=1.0\textwidth]{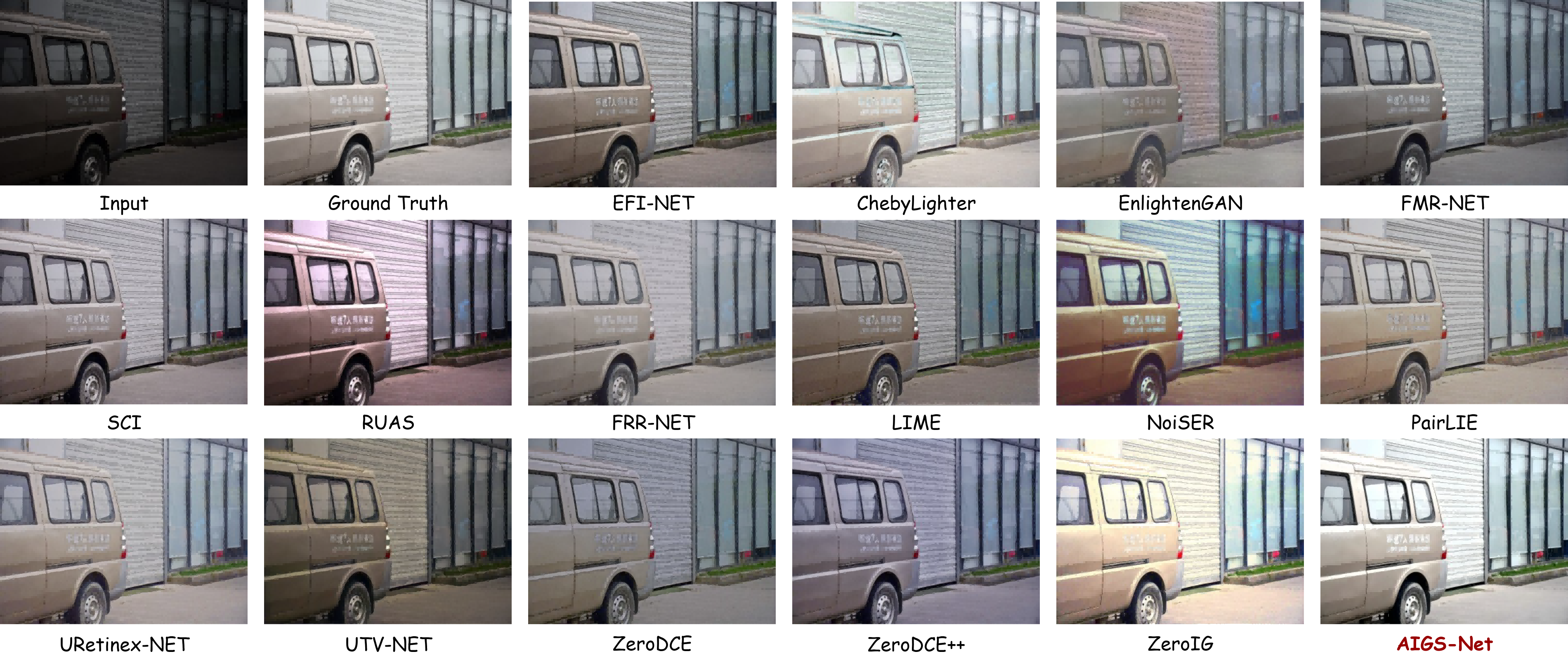}
\caption{Visual comparison between AIGS-Net and SOTA methods on the LSRW-NIKON dataset.}
\label{fig:nikon_visual}
\end{figure*}

\begin{figure*}[!t]
\centering
\includegraphics[width=1.0\textwidth]{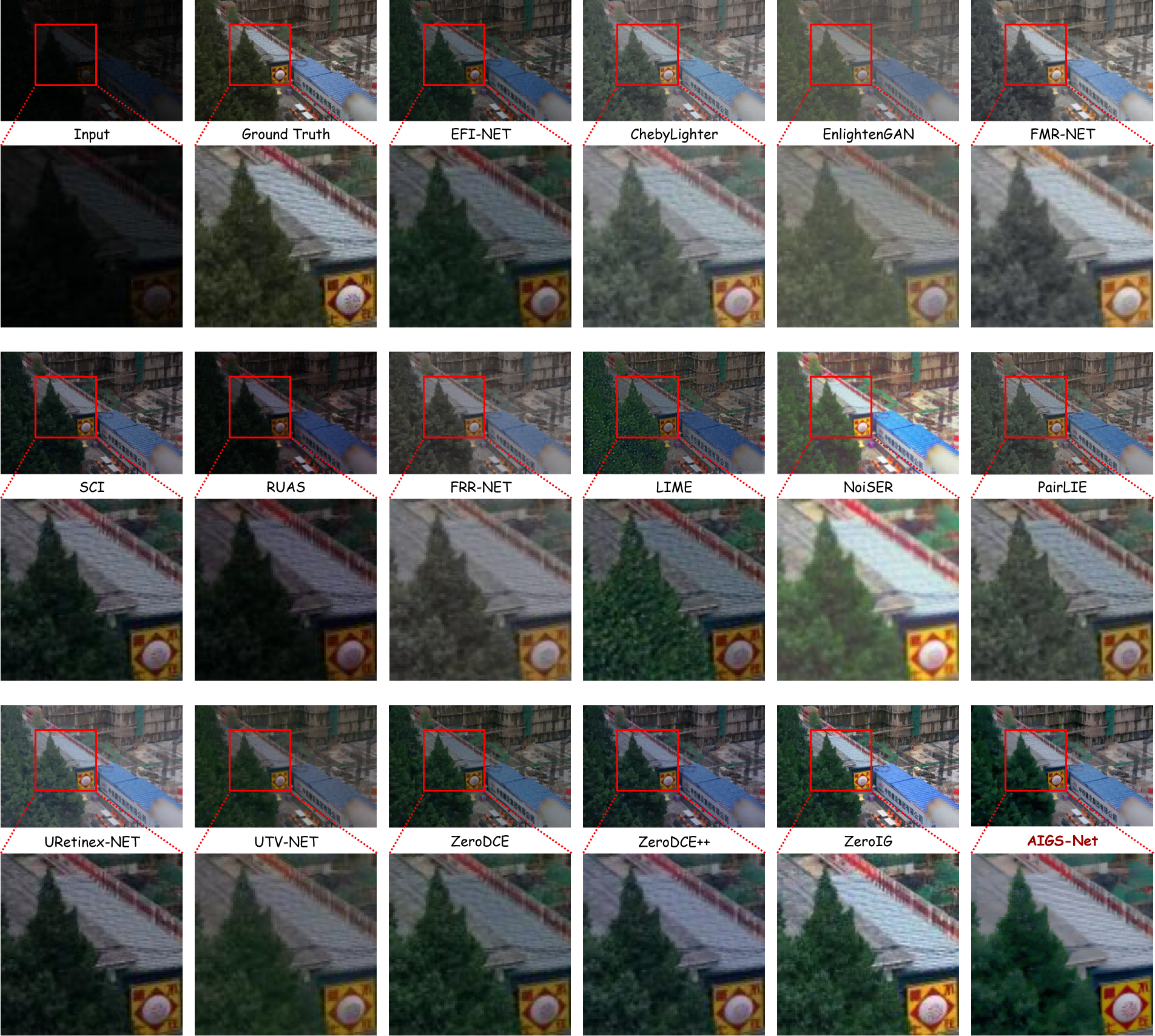}
\caption{Detail comparison between AIGS-Net and SOTA methods on the LOL dataset. For each method, the enlarged view of the region marked by the red box is shown.}
\label{fig:detail_visual}
\end{figure*}

\subsection{Loss Functions}
AIGS-Net is trained in a supervised manner, with reconstruction fidelity, structure preservation, edge constraints, illumination and gain smoothness, color consistency, and input-adaptive diversity constraints jointly considered. Let the normal-light reference image be $I_{gt}$ and the network output be $\hat{I}$.

First, the Smooth $\mathcal{L}_1$ loss is used to constrain the pixel-level difference between the enhanced result and the reference image. Compared with the standard $\mathcal{L}_1$ or $\mathcal{L}_2$ loss, Smooth $\mathcal{L}_1$ imposes a quadratic penalty in small-residual regions and approximates a linear penalty in large-residual regions, making it more robust to outlier pixels:
\begin{equation}
\begin{gathered}
\mathcal{L}_{rec}
=\frac{1}{N}\sum_p \rho\left(\hat{I}_p-I_{gt,p}\right),\\
\rho(x)=
\begin{cases}
0.5x^2, & |x|<1,\\
|x|-0.5, & |x|\geq1 .
\end{cases}
\end{gathered}
\label{eq:rec_loss}
\end{equation}
To preserve local structure and perceptual quality, the SSIM loss is introduced. In the implementation, a Gaussian window with a size of 11 and a standard deviation of 1.5 is used to compute local mean, variance, and covariance:
\begin{equation}
\begin{gathered}
\operatorname{SSIM}_p(A,B)
=\frac{(2\mu_A\mu_B+C_1)(2\sigma_{AB}+C_2)}
{(\mu_A^2+\mu_B^2+C_1)(\sigma_A^2+\sigma_B^2+C_2)},\\
\mathcal{L}_{ssim}
=1-\frac{1}{N}\sum_p \operatorname{SSIM}_p(\hat{I},I_{gt}) .
\end{gathered}
\label{eq:ssim_loss}
\end{equation}
Low-light enhancement should also avoid excessive edge smoothing. Therefore, the Sobel gradient magnitude is computed on the luminance channel, and the edge responses of the output image and the reference image are constrained to be consistent:
\begin{equation}
\begin{gathered}
E(A)=\sqrt{(S_x*Y(A))^2+(S_y*Y(A))^2+10^{-8}},\\
\mathcal{L}_{edge}=\left\|E(\hat{I})-E(I_{gt})\right\|_1 .
\end{gathered}
\label{eq:edge_loss}
\end{equation}
For the input-adaptive illumination field, a spatial smoothness constraint is imposed so that it mainly describes slowly varying illumination rather than high-frequency textures:
\begin{equation}
\mathcal{L}_{illum}=\frac{1}{N_x}\sum_p(\partial_xL(p))^2+\frac{1}{N_y}\sum_p(\partial_yL(p))^2 .
\label{eq:illum_loss}
\end{equation}
For the gain field, both spatial smoothness and cross-channel consistency are constrained. The spatial smoothness term prevents local gain discontinuities from producing color blocks or artifacts, while the cross-channel variance term suppresses abnormal gain in a specific color channel, thereby reducing color casts:
\begin{equation}
\begin{aligned}
\mathcal{L}_{gain}
&=\|\partial_x\Gamma\|_1+\|\partial_y\Gamma\|_1\\
&\quad+\frac{0.5}{N}\sum_{p\in\Omega}
\operatorname{Var}_{c\in\{R,G,B\}}(\Gamma_c(p)) .
\end{aligned}
\label{eq:gain_loss}
\end{equation}
The color consistency loss adopts an angular constraint between RGB color vectors. This paper aligns the output image with the normal-light reference image $I_{gt}$ rather than with the low-light input $I_l$, because the low-light input itself may contain sensor-induced color bias:
\begin{equation}
\begin{gathered}
\mathbf{u}_A(p)=\frac{A(p)+\varepsilon}{\|A(p)+\varepsilon\|_2},\\
\mathcal{L}_{color}
=\frac{1}{N}\sum_{p\in\Omega}
\left(1-\mathbf{u}_{\hat{I}}(p)^\mathrm{T}\mathbf{u}_{I_{gt}}(p)\right) .
\end{gathered}
\label{eq:color_loss}
\end{equation}
In addition, to prevent the input-adaptive illumination field from degenerating into an approximately identical static map for all images, an illumination diversity constraint is introduced. Specifically, the illumination field of each image is adaptively pooled to $16\times16$, flattened, centered, and normalized. Finally, the average cosine similarity between different samples within a batch is minimized:
\begin{equation}
\begin{gathered}
v_i=\operatorname{Flatten}\!\left(\operatorname{Pool}_{16\times16}(L_i)\right),
\tilde{v}_i=\frac{v_i-\operatorname{mean}(v_i)}
{\|v_i-\operatorname{mean}(v_i)\|_2},\\
\mathcal{L}_{div}
=\frac{1}{B(B-1)}\sum_{i\neq j}\tilde{v}_i^\mathrm{T}\tilde{v}_j .
\end{gathered}
\label{eq:div_loss}
\end{equation}
Finally, the total loss is defined as the weighted sum of all loss terms:
\begin{equation}
\begin{aligned}
\mathcal{L}
&=\mathcal{L}_{rec}+0.1\mathcal{L}_{ssim}+0.5\mathcal{L}_{edge}+0.1\mathcal{L}_{illum}\\
&\quad+0.05\mathcal{L}_{gain}+0.2\mathcal{L}_{color}+0.1\mathcal{L}_{div} .
\end{aligned}
\label{eq:total_loss}
\end{equation}
The above loss design reflects the overall principle of the proposed method. Pixel reconstruction and SSIM ensure global fidelity; edge loss preserves structural details; illumination and gain smoothness maintain the stability of the physical priors; cross-channel gain consistency and target color-angle loss suppress color casts; and the illumination diversity term encourages the 2DGS illumination field to remain input-adaptive.

\section{Experiments}
\subsection{Experimental Setup}
\textbf{Dataset.} To systematically evaluate the effectiveness and generalization of the proposed AIGS-Net for low-light image enhancement, this work uses two widely recognized benchmark datasets: the LOL and LSRW datasets~\cite{ref26,ref27}. LOL is the first publicly available paired dataset carefully constructed for low-light enhancement, covering paired low-light and normal-light images from both synthetic and real scenarios. LSRW is the first large-scale paired dataset for real-world scenarios, comprising two subsets captured independently with a Huawei P40 Pro smartphone and a Nikon D7500 digital single-lens reflex camera. Since this paper focuses on lightweight physical-prior modeling and accurate spatially varying illumination compensation, the paired relationship between low-light and normal-light images is strictly preserved during training. Each low-light input and its corresponding normal-light reference are fed into the network in synchrony to establish an interpretable, pixel-level, supervised mapping. All samples are split into training and test sets at a 9:1 ratio, and full-reference metrics are used during testing for objective quantitative evaluation of the enhanced results.

\textbf{Implementation Details.} The entire model is implemented using the PyTorch deep learning framework, and end-to-end training and inference are performed on a single NVIDIA RTX 6000 GPU. During data preprocessing, the original images are first center-cropped to a fixed size of $180\times180$ to construct training samples. In terms of network architecture, the number of Gaussian basis functions in the input-adaptive 2DGS illumination field is set to $K=64$, and the Gaussian centers are initialized as a uniform grid within the normalized coordinate domain $[-0.8,0.8]^2$. The dilation-rate sequence of the multiscale nonlinear contextual encoding module is set to $\{3,5,7\}$, corresponding to receptive fields of $7\times7$, $11\times11$, and $15\times15$, respectively. To ensure the physical stability of the enhancement process, the learnable gain field $\Gamma(x)$ is strictly constrained to $[1.0,12.0]$. The initial logit of the global residual blending weight is set to $-4.0$, so that the network focuses on learning high-brightness enhancement mappings at the early training stage. Meanwhile, to prevent Gamma tone mapping from inducing nonphysical color shifts, the tone curve is designed as a single-channel shared form, forcing the RGB channels to follow the same nonlinear response.

For optimization, the Adam optimizer is used to jointly update all learnable parameters end-to-end. The initial learning rate is set to $5\times10^{-3}$, the batch size is set to 200, and the model is trained for 120 epochs. The learning-rate schedule adopts a MultiStepLR piecewise decay strategy, where the learning rate is multiplied by a decay factor of $\gamma=0.3$ at the 30th, 60th, 90th, and 110th epochs. This schedule ensures fast convergence of Gaussian parameters in the early stage and fine-grained optimization of adaptive opacity in the later stage. To ensure numerical stability of the alpha-depth compositing mechanism and the Gaussian covariance parameters during long-range gradient backpropagation, norm clipping with a threshold of 1.0 is applied to the gradients across the whole network to suppress gradient explosion.

\textbf{Evaluation Metrics.} To comprehensively evaluate the enhanced images in terms of pixel fidelity, perceptual quality, and visual naturalness, seven widely used metrics for image restoration and enhancement are selected: three full-reference and four no-reference metrics. For datasets with paired ground-truth reference images, three full-reference metrics are used to quantify consistency between the enhanced images and reference images: peak signal-to-noise ratio (PSNR), structural similarity index measure (SSIM), and learned perceptual image patch similarity (LPIPS). PSNR mainly quantifies pixel-level fidelity. SSIM measures image similarity across luminance, contrast, and structure. LPIPS evaluates perceptual similarity by computing the distance between enhanced and reference images in the feature space of a deep neural network, using a VGG backbone in this work.

To evaluate image naturalness, contrast, and information richness in scenarios without reference images, especially in real-world deployment or settings lacking paired supervision, four no-reference metrics are further introduced: natural image quality evaluator (NIQE), lightness order error (LOE), discrete entropy (DE), and enhancement measure evaluation (EME). NIQE quantifies image naturalness by measuring the distance between the enhanced image and a natural-image statistical model. LOE evaluates the preservation of relative lightness order during enhancement, which directly reflects how well the adaptive illumination field of AIGS-Net respects the original spatial structure. DE measures information richness, and EME quantifies the magnitude of local contrast variation.

\begin{table*}[!t]
\centering
\caption{Performance comparison results on the LOL dataset. Red and blue fonts indicate the best and second-best results for each metric, respectively.}
\label{tab:lol_results}
\setlength{\tabcolsep}{6pt}
\begin{tabular}{l|ccccccccc}
\toprule
\tabhead{Method} &
\tabhead{SSIM$\uparrow$} &
\tabhead{PSNR$\uparrow$} &
\tabhead{LPIPS$\downarrow$} &
\tabhead{NIQE$\downarrow$} &
\tabhead{LOE$\downarrow$} &
\tabhead{DE$\uparrow$} &
\tabhead{EME$\uparrow$} &
\tabhead{Params\\(K)$\downarrow$} &
\tabhead{FLOPs\\(G)$\downarrow$} \\
\midrule
ZeroDCE++~\cite{ref6} & 0.72 & 20.81 & 0.17 & 5.16 & 23.19 & 2.05 & 20.76 & 10.6 & 0.33 \\
ZeroDCE~\cite{ref7} & 0.71 & 19.36 & 0.16 & 5.05 & 25.43 & 1.93 & 21.03 & 79.4 & 5.21 \\
SCI~\cite{ref8} & 0.67 & 18.36 & 0.17 & 5.84 & 4.30 & 1.96 & 22.21 & \best{0.3} & \second{0.0619} \\
RUAS~\cite{ref9} & 0.53 & 14.02 & 0.21 & 5.33 & \best{0.35} & 1.46 & \best{29.31} & 1.4 & 0.2813 \\
EnlightenGAN~\cite{ref10} & 0.72 & 17.04 & 0.21 & \second{3.68} & 74.79 & 1.03 & 2.54 & 8636 & 61.01 \\
FMR-NET~\cite{ref11} & 0.82 & 19.16 & 0.18 & 3.97 & 29.02 & 2.42 & 4.63 & 196.8 & 102.8 \\
LIME~\cite{ref5} & 0.61 & 17.18 & 0.26 & 5.23 & 93.73 & 1.94 & 23.90 & N/A & N/A \\
FRR-NET~\cite{ref12} & 0.82 & \second{23.63} & 0.18 & 5.29 & 29.06 & 1.91 & 5.51 & 12.21 & 0.216 \\
UTV-NET~\cite{ref13} & 0.81 & 18.83 & 0.13 & 4.36 & 18.79 & 1.84 & 7.22 & 7745 & 58.29 \\
ChebyLighter~\cite{ref14} & 0.82 & 20.43 & 0.10 & 3.98 & 16.32 & 2.31 & 5.85 & 73 & 17.25 \\
EFI-NET~\cite{ref15} & 0.73 & 16.74 & 0.16 & 4.12 & 32.06 & 1.61 & 8.00 & 129.2 & 9.38 \\
PairLIE~\cite{ref16} & 0.69 & 22.74 & 0.20 & 5.59 & 36.27 & 2.22 & 11.94 & 34.18 & 22.35 \\
NoiSER~\cite{ref17} & 0.70 & 15.72 & 0.39 & 3.94 & 47.50 & 2.05 & 3.04 & 1.763 & 8.62 \\
URetinex-NET~\cite{ref18} & \second{0.87} & 22.08 & \second{0.09} & \best{3.60} & 34.33 & 2.27 & 5.95 & 838.3 & 136.01 \\
ZeroIG~\cite{ref19} & 0.60 & 18.18 & 0.21 & 6.00 & 14.08 & \best{2.65} & 22.21 & 123.63 & 118.73 \\
AIGS-Net & \best{0.91} & \best{23.98} & \best{0.07} & 3.71 & \second{7.33} & \second{2.62} & \second{26.46} & \second{0.44} & \best{0.0413} \\
\bottomrule
\end{tabular}
\end{table*}

\begin{table*}[!t]
\centering
\caption{Performance comparison results on the LSRW-HUAWEI dataset. Red and blue fonts indicate the best and second-best results for each metric, respectively.}
\label{tab:huawei_results}
\setlength{\tabcolsep}{6pt}
\begin{tabular}{l|ccccccccc}
\toprule
\tabhead{Method} &
\tabhead{SSIM$\uparrow$} &
\tabhead{PSNR$\uparrow$} &
\tabhead{LPIPS$\downarrow$} &
\tabhead{NIQE$\downarrow$} &
\tabhead{LOE$\downarrow$} &
\tabhead{DE$\uparrow$} &
\tabhead{EME$\uparrow$} &
\tabhead{Params\\(K)$\downarrow$} &
\tabhead{FLOPs\\(G)$\downarrow$} \\
\midrule
ZeroDCE++~\cite{ref6} & 0.66 & 15.86 & 0.25 & 3.47 & 8.85 & 2.28 & 32.98 & 10.6 & 0.33 \\
ZeroDCE~\cite{ref7} & 0.65 & 15.59 & \second{0.22} & 3.48 & 15.54 & 2.23 & 32.62 & 79.4 & 5.21 \\
SCI~\cite{ref8} & 0.59 & 13.80 & 0.25 & 3.43 & 5.21 & 2.07 & 34.14 & \best{0.3} & \second{0.0619} \\
RUAS~\cite{ref9} & 0.49 & 11.37 & 0.35 & 3.75 & \best{0.37} & 1.42 & \second{35.28} & 1.4 & 0.2813 \\
EnlightenGAN~\cite{ref10} & 0.73 & 18.70 & 0.27 & 3.41 & 44.54 & 1.79 & 2.21 & 8636 & 61.01 \\
FMR-NET~\cite{ref11} & \second{0.80} & 19.79 & 0.31 & 4.39 & 29.43 & 2.69 & 3.51 & 196.8 & 102.8 \\
LIME~\cite{ref5} & 0.54 & 14.72 & 0.36 & 4.44 & 155.17 & 2.34 & \best{39.14} & N/A & N/A \\
FRR-NET~\cite{ref12} & 0.74 & \second{20.31} & 0.30 & 4.30 & 13.76 & 2.43 & 5.25 & 12.21 & 0.216 \\
UTV-NET~\cite{ref13} & 0.73 & 18.75 & 0.25 & \best{3.07} & 18.18 & 2.53 & 6.16 & 7745 & 58.29 \\
ChebyLighter~\cite{ref14} & 0.76 & 20.30 & 0.28 & 3.17 & 10.63 & 2.51 & 4.46 & 73 & 17.25 \\
EFI-NET~\cite{ref15} & 0.67 & 13.97 & 0.26 & 3.75 & 19.68 & 1.83 & 6.87 & 129.2 & 9.38 \\
PairLIE~\cite{ref16} & 0.77 & 17.24 & 0.27 & 3.49 & 15.06 & 2.07 & 5.32 & 34.18 & 22.35 \\
NoiSER~\cite{ref17} & 0.76 & 16.08 & 0.46 & 4.38 & 49.04 & 2.29 & 2.09 & 1.763 & 8.62 \\
URetinex-NET~\cite{ref18} & 0.73 & 20.27 & 0.30 & 3.15 & 18.04 & \second{2.70} & 5.23 & 838.3 & 136.01 \\
ZeroIG~\cite{ref19} & 0.65 & 18.29 & 0.23 & 3.82 & \second{4.98} & \best{2.72} & 34.70 & 123.63 & 118.73 \\
AIGS-Net & \best{0.84} & \best{21.60} & \best{0.20} & \second{3.12} & 9.73 & 2.67 & 34.03 & \second{0.44} & \best{0.0413} \\
\bottomrule
\end{tabular}
\end{table*}

\subsection{Performance Comparison}

To comprehensively evaluate the performance of AIGS-Net, extensive qualitative and quantitative comparisons are conducted on the LOL, LSRW-HUAWEI, and LSRW-NIKON datasets against state-of-the-art low-light image enhancement methods~\cite{ref5,ref6,ref7,ref8,ref9,ref10,ref11,ref12,ref13,ref14,ref15,ref16,ref17,ref18,ref19}. The visual quality of qualitative comparisons is shown in Figs.~\ref{fig:lol_visual}--\ref{fig:detail_visual}, and the objective quantitative results are summarized in Tables~\ref{tab:lol_results}--\ref{tab:nikon_results}. In the quantitative evaluation, PSNR, SSIM, and LPIPS are adopted as full-reference metrics, while NIQE, LOE, DE, and EME are used as auxiliary no-reference metrics.

As shown in Fig.~\ref{fig:lol_visual}, different methods exhibit clear visual differences in this classic low-light scene. EFI-NET, SCI, RUAS, ZeroDCE, and ZeroDCE++ still produce overall dark results, with insufficient recovery of dark-region textures and seat details. ChebyLighter, EnlightenGAN, and FRR-NET suffer from haze-like whitening and weak image layering. LIME and ZeroIG introduce noise and color casts, while NoiSER causes overexposure and saturation. In contrast, AIGS-Net provides moderate brightness enhancement, sharper structural edges, and more natural colors, producing results closer to the GT image.

As shown in Fig.~\ref{fig:nikon_visual}, clear differences are also evident in the low-light truck scene. EFI-NET and FMR-NET provide limited brightness improvement, leaving the truck body and rolling-shutter details relatively dark. SCI, PairLIE, and URetinex-NET appear slightly gray and dim. ChebyLighter, EnlightenGAN, and FRR-NET show whitening, haze, and contrast degradation. RUAS, NoiSER, and ZeroIG suffer from color casts, overexposure, or saturation distortion. In contrast, AIGS-Net produces appropriate brightness, clear edge textures, and natural colors, yielding results closer to the GT image.

As shown in Fig.~\ref{fig:detail_visual}, the red boxes and enlarged regions further illustrate local detail differences. EFI-NET, SCI, RUAS, and ZeroDCE remain dark after enhancement, and the roof textures and tree contours remain insufficiently clear. ChebyLighter, EnlightenGAN, FRR-NET, and PairLIE exhibit haze-like whitening and reduced contrast in detail. LIME introduces obvious noise, while NoiSER and ZeroIG produce overexposure, color casts, and structural distortion. In contrast, AIGS-Net better restores roof edges, tree textures, and signboard details, with natural brightness and color closer to those of the GT image.

In evaluations across the three datasets, AIGS-Net achieves the best results on PSNR and SSIM, two mainstream full-reference metrics, with only 440 learnable parameters. It also achieves the lowest FLOPs and ranks among the top three on most other evaluation metrics. Considering its lightweight design, the evaluation results demonstrate that AIGS-Net achieves strong enhancement performance while maintaining extremely high computational efficiency, showing a clear advantage over various SOTA methods in the comprehensive comparison.

\begin{table*}[!t]
\centering
\caption{Performance comparison results on the LSRW-NIKON dataset. Red and blue fonts indicate the best and second-best results for each metric, respectively.}
\label{tab:nikon_results}
\setlength{\tabcolsep}{6pt}
\begin{tabular}{l|ccccccccc}
\toprule
\tabhead{Method} &
\tabhead{SSIM$\uparrow$} &
\tabhead{PSNR$\uparrow$} &
\tabhead{LPIPS$\downarrow$} &
\tabhead{NIQE$\downarrow$} &
\tabhead{LOE$\downarrow$} &
\tabhead{DE$\uparrow$} &
\tabhead{EME$\uparrow$} &
\tabhead{Params\\(K)$\downarrow$} &
\tabhead{FLOPs\\(G)$\downarrow$} \\
\midrule
ZeroDCE++~\cite{ref6} & 0.79 & 16.02 & 0.20 & 3.44 & 75.24 & 0.76 & 6.37 & 10.6 & 0.33 \\
ZeroDCE~\cite{ref7} & 0.76 & 14.87 & 0.18 & 3.31 & 73.95 & 0.39 & 6.33 & 79.4 & 5.21 \\
SCI~\cite{ref8} & 0.70 & 16.82 & 0.20 & 3.26 & 50.63 & 1.05 & 7.61 & \best{0.3} & \second{0.0619} \\
RUAS~\cite{ref9} & 0.68 & 14.03 & 0.26 & 3.33 & \best{0.99} & \second{1.30} & \best{8.26} & 1.4 & 0.2813 \\
EnlightenGAN~\cite{ref10} & 0.79 & 16.46 & 0.20 & 3.25 & 136.37 & 0.37 & 2.77 & 8636 & 61.01 \\
FMR-NET~\cite{ref11} & 0.80 & 15.24 & 0.19 & 3.79 & 13.32 & 0.87 & 3.78 & 196.8 & 102.8 \\
LIME~\cite{ref5} & 0.66 & 13.95 & 0.25 & 3.25 & 74.58 & 0.85 & 8.06 & N/A & N/A \\
FRR-NET~\cite{ref12} & \second{0.83} & \second{19.83} & 0.16 & 4.08 & 17.55 & 0.52 & 2.92 & 12.21 & 0.216 \\
UTV-NET~\cite{ref13} & 0.70 & 11.32 & 0.22 & 3.64 & 21.92 & 0.70 & 5.88 & 7745 & 58.29 \\
ChebyLighter~\cite{ref14} & 0.74 & 16.10 & 0.21 & \best{3.14} & 83.96 & 0.77 & 3.93 & 73 & 17.25 \\
EFI-NET~\cite{ref15} & 0.74 & 15.47 & 0.20 & \second{3.15} & 45.22 & 0.98 & 5.41 & 129.2 & 9.38 \\
PairLIE~\cite{ref16} & 0.72 & 11.05 & 0.21 & 3.66 & \second{9.26} & 0.41 & 7.25 & 34.18 & 22.35 \\
NoiSER~\cite{ref17} & 0.76 & 15.31 & 0.39 & 3.77 & 29.42 & 1.00 & 3.25 & 1.763 & 8.62 \\
URetinex-NET~\cite{ref18} & 0.82 & 19.74 & \second{0.14} & 3.20 & 36.36 & 0.60 & 3.60 & 838.3 & 136.01 \\
ZeroIG~\cite{ref19} & 0.63 & 14.13 & 0.26 & 3.37 & 48.12 & 0.62 & 6.20 & 123.63 & 118.73 \\
AIGS-Net & \best{0.87} & \best{20.41} & \best{0.13} & 3.21 & 12.05 & \best{1.40} & \second{8.09} & \second{0.44} & \best{0.0413} \\
\bottomrule
\end{tabular}
\end{table*}

\begin{table}[!tbp]
\centering
\caption{Quantitative comparison of different illumination-field modeling strategies. Red and blue fonts indicate the best and second-best results for each metric, respectively.}
\label{tab:illumination_ablation}
\setlength{\tabcolsep}{4pt}
\begin{tabular}{clccc}
\toprule
Model & Representation Type & SSIM$\uparrow$ & PSNR$\uparrow$ & LPIPS$\downarrow$ \\
\midrule
A & Convolution only & 0.74 & 19.61 & 0.18 \\
B & Static 2DGS & 0.80 & 21.08 & 0.15 \\
C & Adaptive 2DGS & \second{0.89} & \second{21.97} & \second{0.13} \\
D & Ours & \best{0.91} & \best{23.98} & \best{0.07} \\
\bottomrule
\end{tabular}
\end{table}

\subsection{Ablation Study}
To further investigate the rationality and contribution of each core component and physical-prior design in AIGS-Net, comprehensive ablation experiments are conducted on the LOL dataset. To ensure fair evaluation, all variant models adopt the same training strategy and hyperparameter settings. The evaluation metrics include PSNR, SSIM, LPIPS, and the number of learnable parameters.

\textbf{Effectiveness of the Input-Adaptive 2DGS Illumination Field.} Illumination-field modeling is the core innovation of this paper. To address the limitations of early methods that use static priors or absolute-luminance modulation, the following variants are designed for comparison: 1) Model A (w/o 2DGS), where the whole Gaussian Splatting module is removed and a pure convolutional branch is used to directly predict the illumination gain; 2) Model B (Static 2DGS), where fixed Gaussian opacity $\alpha$ is adopted, so the illumination field does not vary with the input image and becomes a static mapping; 3) Model C (Adaptive 2DGS), where the absolute luminance $Y(\mu_k)$ at each Gaussian center is used to modulate $\alpha$; and 4) Model D (Ours), where $\alpha$ is dynamically modulated based on relative luminance statistics normalized by $\bar{Y}$ and $\sigma_Y$.

\begin{table}[!tbp]
\centering
\caption{Ablation results of the zero-parameter contextual encoding module. Red and blue fonts indicate the best and second-best results for each metric, respectively.}
\label{tab:context_ablation}
\setlength{\tabcolsep}{2pt}
\begin{tabular}{ccccc}
\toprule
\tabhead{Linear Mean} & \tabhead{Linear Contrast} & \tabhead{SSIM$\uparrow$} & \tabhead{PSNR$\uparrow$} & \tabhead{\shortstack{Additional \\ Params}} \\
\midrule
$\times$ & $\times$ & 0.77 & 21.02 & 0 \\
$\checkmark$ & $\times$ & 0.79 & 21.37 & 0 \\
$\times$ & $\checkmark$ & \second{0.84} & \second{21.94} & 0 \\
$\checkmark$ & $\checkmark$ & \best{0.88} & \best{23.09} & 0 \\
\bottomrule
\end{tabular}
\end{table}

As shown in Table~\ref{tab:illumination_ablation}, Model A, which relies only on a pure convolutional structure with very few parameters, has extremely limited spatial representation capacity. After 2DGS is introduced, Model B achieves improved performance. However, because the mapping is static, it cannot handle abrupt spatial luminance changes across different scenes. Model C, based on absolute-luminance adaptation, has a critical limitation: under extremely low-light conditions, the absolute luminance of all pixels is very low, leading to homogeneous predicted illumination fields. In contrast, the proposed Model D overcomes this limitation by using relative luminance and achieves a PSNR improvement of more than 4.37 dB over Model A.

\begin{table}[!tbp]
\centering
\caption{Comparison results of the physical degradation restoration module. Red and blue fonts indicate the best and second-best results for each metric, respectively.}
\label{tab:degradation_ablation}
\setlength{\tabcolsep}{2pt}
\begin{tabular}{lccp{0.36\columnwidth}}
\toprule
Strategy & SSIM$\uparrow$ & PSNR$\uparrow$ & Visual Observation \\
\midrule
w/o Noise Mask & 0.83 & 22.08 & Severe noise in dark areas \\
Multi-Channel Gamma & \second{0.85} & \second{22.52} & Yellow/blue shift observed \\
Ours & \best{0.88} & \best{23.09} & Preserved white balance \\
\bottomrule
\end{tabular}
\end{table}

\begin{table}[!tbp]
\centering
\caption{Effectiveness verification of the loss functions. Red and blue fonts indicate the best and second-best results for each metric, respectively.}
\label{tab:loss_ablation}
\setlength{\tabcolsep}{4pt}
\begin{tabular}{ccccccc}
\toprule
Basic Loss & $\mathcal{L}_{illum}$ & $\mathcal{L}_{div}$ & $\mathcal{L}_{gain}$ & $\mathcal{L}_{color}$ & SSIM$\uparrow$ & PSNR$\uparrow$ \\
\midrule
$\checkmark$ & $\times$ & $\times$ & $\times$ & $\times$ & 0.74 & 20.59 \\
$\checkmark$ & $\checkmark$ & $\times$ & $\times$ & $\times$ & 0.79 & 21.33 \\
$\checkmark$ & $\checkmark$ & $\checkmark$ & $\times$ & $\times$ & \second{0.86} & 22.57 \\
$\checkmark$ & $\checkmark$ & $\checkmark$ & $\checkmark$ & $\times$ & \second{0.86} & \second{22.79} \\
$\checkmark$ & $\checkmark$ & $\checkmark$ & $\checkmark$ & $\checkmark$ & \best{0.88} & \best{23.09} \\
\bottomrule
\end{tabular}

\end{table}

\begin{figure}
\centering
\includegraphics[width=\columnwidth]{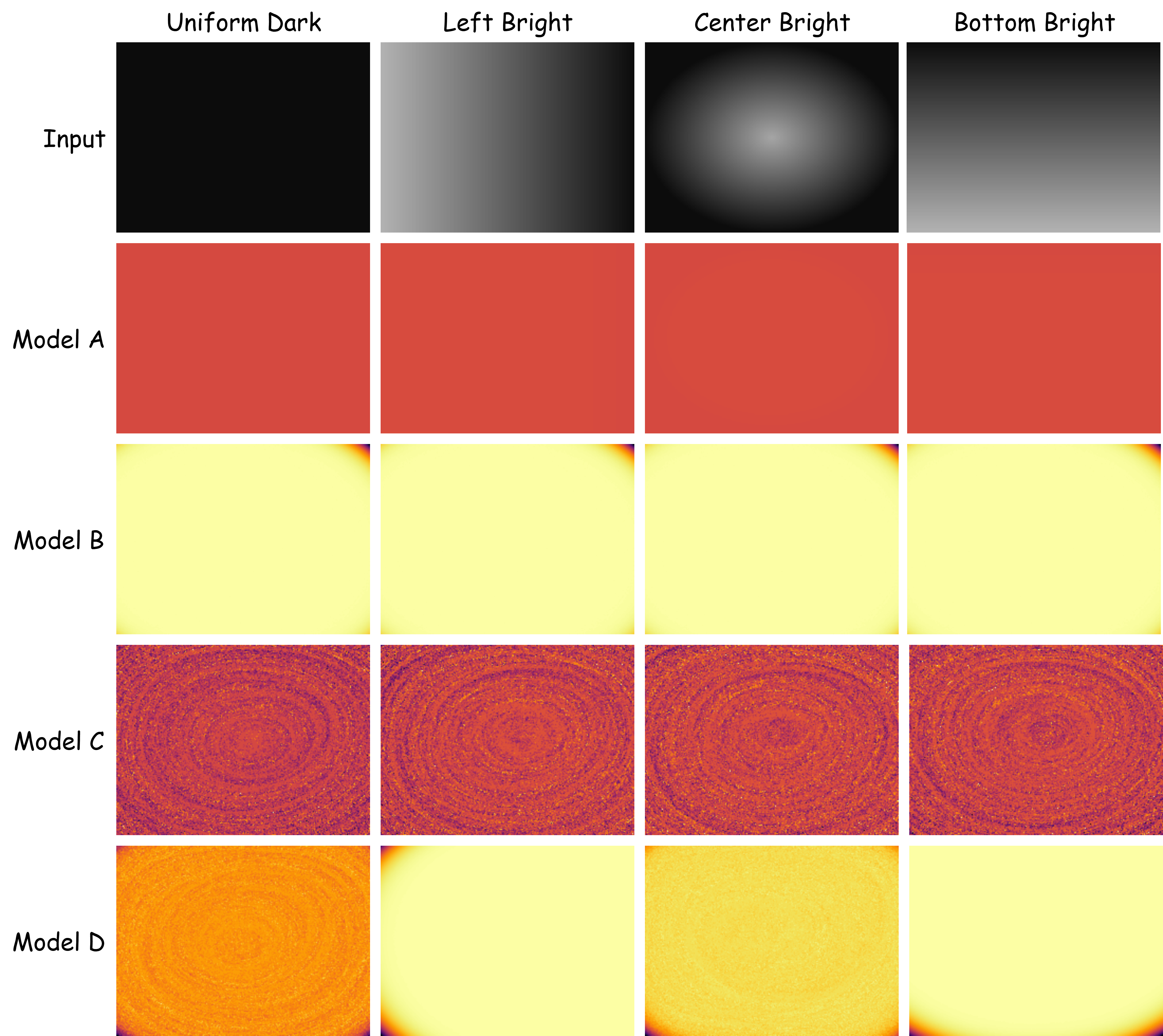}
\caption{Input-adaptive responses of different 2DGS illumination-field modeling strategies. The first row shows four synthetic low-light inputs, and the remaining rows show the illumination fields generated by Models A--D. The color follows a Pantone-inspired soft gradient, indicating responses from low to high. Compared with static and absolute-luminance modulation, Model D produces more distinct input-dependent illumination fields that reflect the relative distribution of dark regions, consistent with the best quantitative results in Table~\ref{tab:illumination_ablation}.}
\label{fig:adaptive_response}
\end{figure}

As shown in Fig.~\ref{fig:adaptive_response}, four synthetic low-light inputs with different spatial luminance distributions are generated to verify the illumination field input-adaptive capability. Model A relies only on convolution and generates a relatively uniform illumination field. Model B adopts static 2DGS and produces nearly identical responses for different inputs. Model C introduces absolute-luminance modulation, but its ability to distinguish different luminance structures remains limited. In contrast, Model D dynamically modulates Gaussian opacity based on relative luminance statistics and can generate differentiated illumination fields according to the dark-region distribution of each input.

\textbf{Ablation on Zero-Parameter Nonlinear Contextual Encoding.} To verify the effectiveness of pure shift aggregation in guiding illumination gain, the contextual encoding module is isolated and evaluated using four variants: 1) w/o Context, where the contextual encoding module is removed and the 2DGS result is directly concatenated with the input for gain prediction; 2) w/ Linear Mean Only, where only the local mean branch is retained to provide low-frequency structural guidance; 3) w/ Nonlinear Contrast Only, where only the local nonlinear contrast branch is retained to provide high-frequency texture guidance; and 4) Ours, where the mean feature is used as the gated value and the contrast feature is used as the gating weight.

As shown in Table~\ref{tab:context_ablation}, this module provides a 2.07 dB gain without introducing additional learnable parameters. Using only the mean branch improves low-frequency illumination patterns, while using only the contrast branch better preserves edges. The complete design uses contrast as the gating signal to activate the mean features, which successfully 
suppresses overexposure in flat regions and detail loss in textured regions during brightening.

\textbf{Physical Degradation Restoration and Color-Locking Mechanism.} To address noise degradation and color-bias preservation during low-light amplification, ablation experiments are conducted on the noise-mask estimation and Gamma mapping strategies. Three variants are designed: 1) w/o Noise Mask, where the noise term in the equation is removed; 2) Multi-Channel Gamma, where RGB channels are allowed to learn independent Gamma values; and 3) Ours, where the noise mask is enabled and single-channel Gamma is locked to enforce relative white balance.

As shown in Table~\ref{tab:degradation_ablation}, when the noise mask is removed, the network is forced to reduce the upper bound of the gain to avoid noise amplification, leading to insufficient overall brightness and a sharp decrease in SSIM. When multi-channel Gamma is adopted, the network uses the additional degrees of freedom to overfit the inherent sensor color bias, thereby degrading the color consistency of the enhanced results. The proposed single-channel locked Gamma blocks the pathway through which color bias is preserved at the architectural level.

\textbf{Effect of Physical-Prior Loss Functions.} This paper designs a combined loss function consistent with low-level photometric principles. In this experiment, the proposed specific loss terms are added one by one on the basis of the basic losses $\mathcal{L}_{rec}$, $\mathcal{L}_{ssim}$, and $\mathcal{L}_{edge}$. As shown in Table~\ref{tab:loss_ablation}, each loss term demonstrates its core function. When $\mathcal{L}_{div}$ is introduced, PSNR is improved by 0.3 dB. Without this constraint, 2DGS can easily fall into mode collapse, where all Gaussian basis functions tend to output the same uniform brightness. When $\mathcal{L}_{gain}$ is introduced, PSNR increases by 1.24 dB, confirming the importance of this term. After introducing $\mathcal{L}_{color}$, PSNR further increases by 0.22 dB. More importantly, the anchor for color cosine similarity is set to the target image rather than the color-biased input image, further correcting the inherent bias of low-light sensors.

\section{Conclusion}
This paper proposes AIGS-Net, an ultra-lightweight network for low-light image enhancement. To address the bottleneck between conventional illumination modeling and computational cost, an input-adaptive 2DGS illumination field is constructed. Gaussian opacity is dynamically modulated by relative luminance statistics, enabling accurate spatially varying illumination compensation. Meanwhile, zero-parameter nonlinear multiscale contextual encoding is introduced to efficiently guide structure-aware gain estimation. Single-channel Gamma mapping and cross-channel consistency constraints are further incorporated to physically block the preservation of low-light noise and color casts. Experimental results show that AIGS-Net improves detail recovery and color fidelity with only approximately 440 learnable parameters. It reaches a favorable trade-off between visual quality and extreme inference efficiency, providing a promising solution for real-time vision deployment on resource-constrained devices.

\end{document}